\documentclass[letterpaper, 10 pt, conference]{ieeeconf}  

\IEEEoverridecommandlockouts                              

\usepackage{cite}
\usepackage{amsmath,amssymb,amsfonts}
\usepackage{algorithmic}
\usepackage{graphicx}
\usepackage{textcomp}
\usepackage{hyperref}
\usepackage{subfigure}

\usepackage{xcolor}

\begin{document}

\title{A Human-friendly Verbal Communication Platform for Multi-Robot Systems: Design and Principles}

\author{Christopher Carr$^{1}$, Peng Wang$^{*,1}$, Shenglin Wang$^{2}$
\thanks{*Correspondence to Peng Wang: p.wang@mmu.ac.uk}
\thanks{$^{1}$Christopher Carr and Peng Wang are with the Department of Mathematics and Computing,
        Manchester Metropolitan University, M15 6BH, United Kingdom
        {\tt\small p.wang@mmu.ac.uk, christopher.carr@stu.mmu.ac.uk}}%
\thanks{$^{2}$Shenglin Wang is with the Department of Automatic Control and Systems Engineering, The University of Sheffield,
        Sheffield, S10 2TN, UK
        {\tt\small swang119@sheffield.ac.uk}
        }%
}

\maketitle

\begin{abstract}

While multi-robot systems have been broadly researched and deployed, their success is built chiefly upon the dependency on network infrastructures, whether wired or wireless. Aiming at the first steps toward de-coupling the application of multi-robot systems from the reliance on network infrastructures, this paper proposes a human-friendly verbal communication platform for multi-robot systems, following the deliberately designed principles of being adaptable, transparent, and secure. The platform is network independent and is subsequently capable of functioning in network infrastructure lacking environments from underwater to planet explorations. A series of experiments were conducted to demonstrate the platform's capability in multi-robot systems communication and task coordination, showing its potential in infrastructure-free applications. To benefit the community, we have made the codes open source at \href{https://github.com/jynxmagic/MSc_AI_project}{GitHub}.

\end{abstract}

\section{Introduction}

The continuous development of robotic technologies along with the significant advancements achieved by Artificial Intelligence (AI) techniques such as Natural Language Processing (NLP) and Computer Vision (CV) have driven the massive deployments of robots in sectors like manufacturing, warehouse, and healthcare. Recently, being still in its infancy, multi-robot systems (MRS) have attracted attentions as they excel a single robot in terms of robustness and efficiency in task execution, or can deliver tasks that are impossible for a single robot~\cite{parker2016multiple}. Despite of the advantages, MRS cannot be simply considered as the generalisation of techniques that are applied in single-robot scenarios~\cite{verma2021multi}, which further stimulates the research and applications of MRS~\cite{parker2016multiple}.


Ever since the introduction of MRS in 1980s~\cite{parker2016multiple}, an array of topics have been investigated ranging from biological inspired MRS techniques, communication, architectures, task allocation, localisation, mapping, exploration, object transport \& manipulation, motion coordination, to re-configurable robots~\cite{arai2002advances}. A few taxonomy paradigms have been proposed aiming at categorised the topics~\cite{verma2021multi,ismail2018survey}. Roughly, there are three features that are widely used for MRS classification: \textbf{1)} Types of robots. MRS can be broadly broken down into two types, i.e., homogeneous and heterogeneous. In a homogeneous MRS, all robots are of equivalent physical structure and follow similar objectives, whereas robots in a heterogeneous MRS can have differing characteristic and objectives~\cite{quinn2001comparison}. \textbf{2)} Control architectures. The control architecture is literally the `brain' of the MRS, and it is crucial for task coordination to deliver collaborative and cooperative jobs. There are mainly three control architectures, i.e., decentralised control (reactive control), centralised control (deliberative control), and hybrid architectures, which is a combination of centralised and decentralised controls. The hybrid control is mostly used as it works in a hierarchical way: in the low level, decentralised control is used to coordinate (percept each other then act based on perceptions) between robots directly while in the high level, centralised control is used to provide the MRS with rich sensor perception and the full environment information for coordination and collaboration. \textbf{3)} Communication mechanisms~\cite{ismail2018survey}. Communication can be broken down into what is known as `explicit' and `implicit' communications. Implicit communication refers to communication which is discovered via the state-change of an environment. For example, if a robot moved a box and another robot discovered the box had been moved, the result is implicit communication. Explicit communication occurs when robots communicate with each other for specific reasons solely with the intent of communication and no environmental state changes~\cite{arai2002advances}. Researchers have discovered that in many cases, explicit communication of even a small amount can lead to drastic improvements within the MRS~\cite{balch1994communication}. 

\begin{figure*}[ht]
    \centering
    \includegraphics[width=0.85\linewidth]{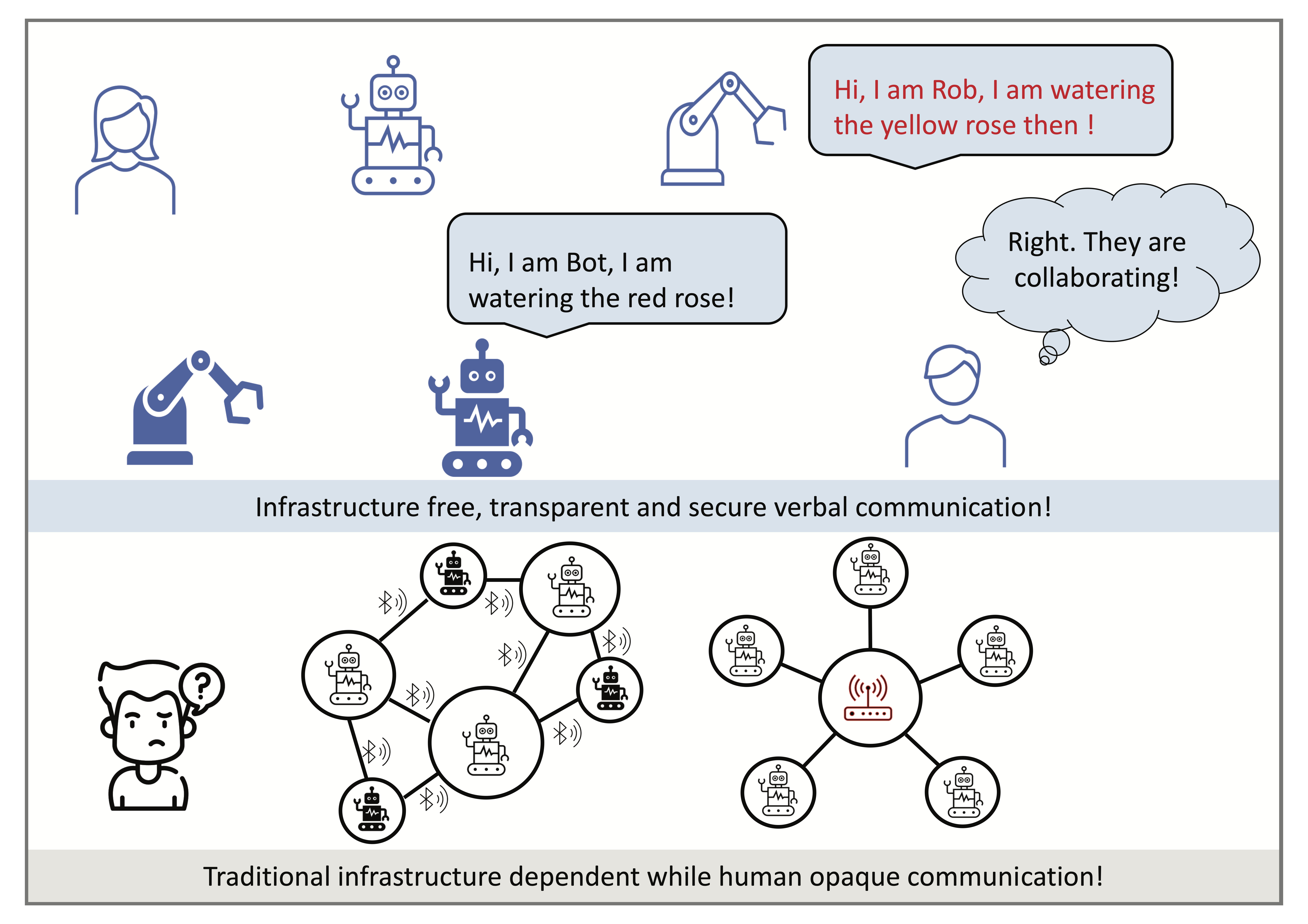}
    \caption{Comparison of the proposed verbal communication platform (top) against traditional bluetooth based decentralised MRS (bottom middle) and Wi-Fi based centralised MRS (bottom right) which reply on infrastructures and the communication is opaque to humans. The proposed verbal communication platform is, in contrast, infrastructure free and transparent and secure to humans.}
    \label{fig:HVCP}
\end{figure*}

Giving it implicit or explicit, current MRS communication relies on certain `infrastructures', from wired to wireless networks. While there is no doubt that Wireless Fidelity (Wi-Fi) network has become the \textit{de facto} facility for MRS communication~\cite{behera2018wireless}, some constraints remain: 1) Wi-Fi routers operating on the 2.4 GHz band, and the distance the signals can reach is affected by blockage and bypass, etc.
Currently, redundant Access Points (APs) are installed to deal with such issues, adding extra costs to the whole system and making the system more infrastructure dependant. 2) Wi-Fi signals can only reach centimetres underwater, which makes any Wi-Fi network based MRS solutions ineligible to underwater applications. 3) The communication within MRS is opaque to humans, which makes it difficult, or at least requires extra efforts to monitor issues such as ethical biases in the communication.

Aiming at dealing with the above issues in MRS communication, this paper proposes a human-friendly verbal communication platform (HVCP). The HVCP refers to the scenario where a robot speaks verbally to another robot, which speaks back through microphone or speakers. Figure \ref{fig:HVCP} shows a comparison of the HVCP and traditional bluetooth and Wi-Fi based communication frameworks. The HVCP will enable MRS to communicate verbally and  network infrastructure free, henceforth possesses the potential of being used in challenging tasks such as planet/underwater explorations. The contributions of this paper include: 1) Robot-to-Robot (R2R) communication is carried out verbally, i.e., robots are verbally talking to each other directly for coordination and collaboration, etc. Compared to Wi-Fi signals, verbal signals can travel as far as 180 metres in air and 2,030 metres underwater, which enables MRS to work in a larger area. 2) No network infrastructure is needed for communication. As robots are talking verbally, network that traditionally meant to connect robots becomes unnecessary. This drastically expand the applications of MRS to scenarios where network is not applicable. 3) The communication is transparent to humans, i.e., humans can monitor the communication at any time and anywhere to avoid biases in communications that would lead to discrimination or biases in task execution stage.

The remainder of the paper is organised as follows. Some related works are introduced in Section II. Section III elaborates on the principles and design of the human-friendly verbal communication platform. Implementation, results and analysis are provided in Section IV, and the paper is concluded in Section V.

\section{Related Works}

Verbal communication is continuously attracting attention from both industry and academia, which as Marin, et al. claims is driven by the desire of developing and achieving human-like interaction between humans and robots~\cite{marin2021verbal}. The benefits of verbal communication between human and robots include but not limited to robot therapy~\cite{melo2019project}, educational robotics~\cite{budiharto2017edurobot}, assistive robotics~\cite{zhou2018new}, etc. Compared to non-verbal communication solutions, verbal communication possesses the potential of bridging the knowledge gap between techniques and non-expert humans, and breaking the ethical barriers to eventually make robots a part of the human society~\cite{kuipers2020perspectives}. In literature, the communication interface through which a human communicates and delivers/extracts data to a robot can either be statically programmed or learned through an AI model. Both of these have their own advantages, with AI models being much more adaptive to circumstances. Among all the techniques related, text-to-speech and speech-to-text are at the heart of verbal communication. Consequently, the challenges in these two areas undermine the capacity of verbal communication. For instance, the capture quality of speech is affected by both the distance from the speaker to the receiver and the ambient noise level~\cite{kumatani2012microphone,marin2021verbal}. When multiple speakers communicate to one or more receiver robots simultaneously, there is the risk of intentions not delivered to the correct destinations due to the lack of speaker identification~\cite{goeldner2015emergence,marin2021verbal}. Even if there is speaker identification mechanisms, the changing of voice along aging process could still hinder the performance of the verbal communication systems~\cite{kennedy2017child,marin2021verbal}. 

It is surprising that while human-to-robot (H2R) verbal communication has been so popular, R2R verbal communication has barely touched. To the best of the authors' knowledge, there is no published solutions for R2R verbal communication. The most popular R2R communication is mostly achieved by Wi-Fi~\cite{chen2020wireless}, Bluetooth~\cite{mahmood2013implementation}, or Ultra-wideband (UWB)~\cite{shule2020uwb} techniques, which are efficient but not human friendly. There is also work that aims at enabling MRS to develop their own language to achieve communication. The issue still is the developed language cannot be understood by humans~\cite{mordatch2018emergence}. One would argue that R2R verbal communication can be naturally solved by mature H2R solutions. The authors agree that there is the possibility, but believe an independently designed human-friendly verbal communication platform following the principles of adaptability, transparency, and security will massively benefit the MRS community. This will be discussed in the next section.

\section{Design Principles and Architecture}

Inspired by the H2R achievements in literature, our vision in designing and implementing the R2R verbal communication platform is such that it is infrastructure (cable or APs) and robot type (heterogeneous or homogeneous) independent, which makes it an adaptable tool for general coordination and collaboration purposes. The long-term ambition to eventually bring MRS based on R2R verbal communication platform into human society also requires it to be secure and transparent to humans. We will detail each of the three aspects as follows.

\subsection{Adaptability}


The development of a verbal communication platform through which the robot communicates requires for the ability of robots to not only encode the information which it is attempting to transfer, but also for the receiving robot to be able to decode that information into useful data. Therefore, there is the need of co-evolution of encoders and decoders~\cite{lipson2007evolutionary}. 



Encoders and decoders for human language of high quality are freely available to most people. The co-evolution of decoders and encoders would be offloaded onto a completely separate area of research, mainly text-to-speech and speech-to-text, which are hot spots in NLP and successful application such as DALL-E is able to locate objects according to natural language commands. With the development of text-to-speech and speech-to-text technologies, one can foresee that robots would be able to communicate heterogeneously, to any other different types of robot, using a verbal communication platform. They would all be using the same definitions and protocols for communication (language), thus allowing a team of many different types of robots and software to communicate for coordination and collaboration, and the usage of verbal signals for communication will massively reduce the dependency of infrastructures and make MRS adaptable!


\subsection{Transparency}
There is no doubt that a verbal communication based MRS can be stand-alone and robots-only. However, the anticipation for robots being -as-a-service is prevailing that it will inevitably play its part in human society. For instance, Iqbal and Riek performed some research on the dynamics of a multi-robot, multi-human heterogeneous team~\cite{iqbal2017coordination}. They discovered that adding a robot to a human team can alter its dynamics drastically. In their work, they based most of what could be done based on what the human team could see the robot team was doing. If those robots were communicating verbally, the human team would have been much more apt at adapting to the robot teams' behaviours. Verbal communication in MRS gives the added advantage of transparency to human operators around them. It will be much easier for such teams (human and robot) to exist together and perform more complex tasks. The added level of transparency not only allows humans to understand what the robot team is doing, but also intervene if the robot team is doing something wrong. This is extremely important to the ethics of AI~\cite{hagendorff2020ethics}. Being as part of AI research, the ethical issues such as discrimination present in AI also present themselves in robotics. The issue of discrimination in robotics can be partly solved by using verbal communication. For instance, the verbal communication platform will enable humans, being operators or just normal humans, will be able to hear inter-process communication as it happens, and are much more likely to spot discrimination as it happens and subsequent actions can be taken to deal with the situation. 

Another advantage of such transparency is the ability to `plug-and-play' human operators into a MRS. Humans, at any intelligence level, can easily join in to the MRS and begin understanding or giving commands to the robots within the team. Along as the robots are communicating verbally, it makes no difference to add a human operator at any point to join in on the task. This is extremely advantageous. Whilst robot teams may not need to take elongated breaks, humans do. There will be periods of over 12 hours where no human operators could be in the mix and the robots are still working (after people finish work). Or, if shift work is in place, another group of human operators would need to come and replace the existing human operators within the team. It would be extremely simple for these humans to add themselves to the robot team and understand what the robots are currently doing.

\subsection{Cybersecurity}
Robots, being traditionally connected to a network with open ports to enable communication, are presenting themselves to multiple cybersecurity attacks. A MRS functioning as such exposes itself to hackers can lead to malicious attacks, and consequences could be unbarable if human is involved in the task. The interests of businesses and people alike need to be protected against hackers by developing mechanisms to stop hackers from gaining unauthorised access to a MRS~\cite{lacava2021cybsersecurity}.

As mentioned earlier, communicating verbally makes a MRS less infrastructure dependent, and almost free of traditional network. This makes cyberattacks on the robots very difficult, as hackers will have to connect to the robots manually (i.e. `physically connect' to the robot). This can be easily tackled by measures such as spotted and reported by humans that are collaborating with the robots. It, however, leaves arguable aspects such as hackers can still control the robots in absence of human operators. In that case, verbal communication between robots could be deliberately encrypted, with each robot containing its own access key to decrypt the communications. Hackers would then have no way of gaining access to the communications, without studying the communications at length. Overall, communication through the sound channel (verbally), rather than a network channel, makes the MRS almost impervious to an attack.


\subsection{Architecture}

\begin{figure}[tbp]
    \centering
    \includegraphics[width=1.0\linewidth]{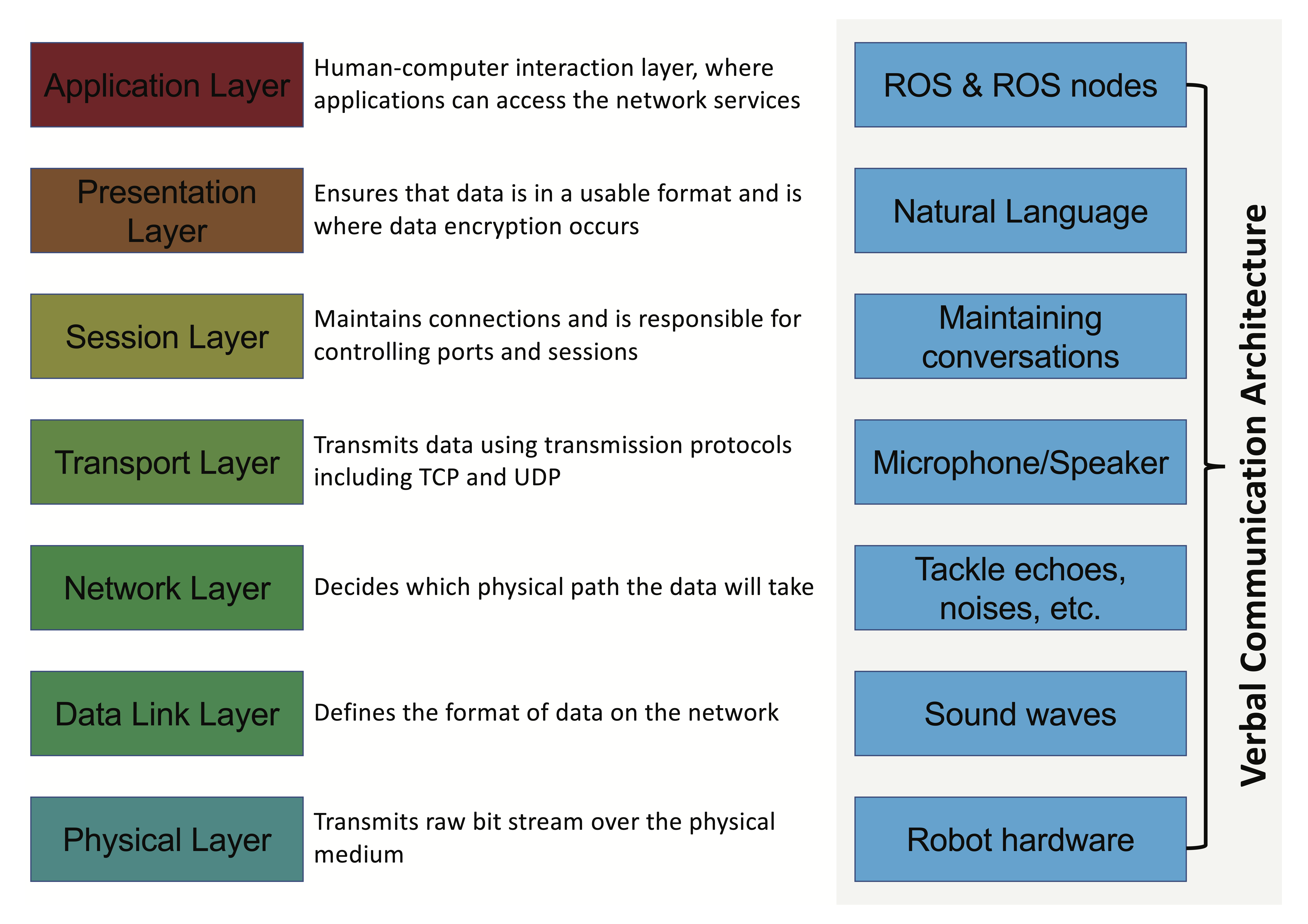}
    \caption{Architecture of the verbal communication platform and its correspondence with the OSI model.}
    \label{fig:architechture}
\end{figure}

Bearing the three principles in mind, the architecture of the proposed verbal communication platform is designed to be as generalised and adaptable as possible. The Open Systems Interconnection (OSI) model is used as a basis for the platform to ensure heterogeneity and usability. In this paper, each OSI model layer as shown in Figure \ref{fig:architechture} is mapped to its functionality in the proposed platform. The OSI physical layer corresponds to the \textit{robot hardware} layer. The Data Link Layer and Network Layer in the OSI model are replaced by verbal signals, indicating the independence of our platform on infrastructures. The Network Layer, however, may need adjustments for long distance verbal communication, or scenarios where there are noises and echoes. The Transportation Layer in the proposed platform is implemented by speakers to either record speech or read out loud text information for communication. The Session Layer, the Presentation layer, and the Application Layer are meant to maintain the conversations among robots, perform text-to-speech and speech-to-text translation, and integrate the platform with ROS to make it usable to the ROS driven robotics community, respectively.

Given the general architecture, a complete communication between any two robots for any task will include three steps, i.e., 1) greeting, 2) task definition, and 3) roles \& responsibilities coordination. The initial greeting was settled to be a template: `Hi, I am X'. This initial greeting serves the purpose of sharing a unique identifier to other robots, so that other robots may contact that robot specifically. After greeting, each of the robots has the unique identifier of all other robots within its vicinity. Also, a unique identifier can be used to contact specific groups of robots, or all robots. For example: `Team A', `Team B', and `All'. 
The next stage within communication is task definitions. The robots can simply share names of the tasks they are completing. This helps to identify robots which are collaborating on a task: `Hi all, I am X, member of Team A, completing task: water the red rose'. These communications can be broken down into separate statements, though it should be noted that all robots are required to specify their name during communications as 

`\textit{Hi}, [robot identifier], \textit{I am X}, [statement]'. 

\noindent
This allows other robots to understand which robot the communication is coming from and what they are doing. It is worth noting that in our implementation, we use a task repository to maintain tasks and the statement of each task is broken down to a series of actions such as \textit{moving forward} to deliver the task.

Robots, along with defining the task, must also define their responsibilities within a task. This mostly relates to collaboratively delivering a task between robots. Assuming two robots are completing the task `water plants', they need to distribute the task between them so that each will water some different plants. To achieve efficient coordination and collaboration, robots are sharing task repositories. It defines specifically what the task is and what the responsibilities of the robots are within that task. For the `water plants' task, the statement looks as follows: `Hi, [team plant], I am X, I will complete task: 1'. Other robots should also assign themselves to responsibilities within the task: `Hello [team plant], I am Y, I will complete task: 2'. Repeating in this fashion, the robots will distribute the tasks between themselves until no tasks remain. Again, it is very important to note that each of the robots knows exactly what the other robots are doing so as to not assign themselves to the same task. 

\begin{figure}[tbp]
    \centering
    \includegraphics[width=1.0\linewidth]{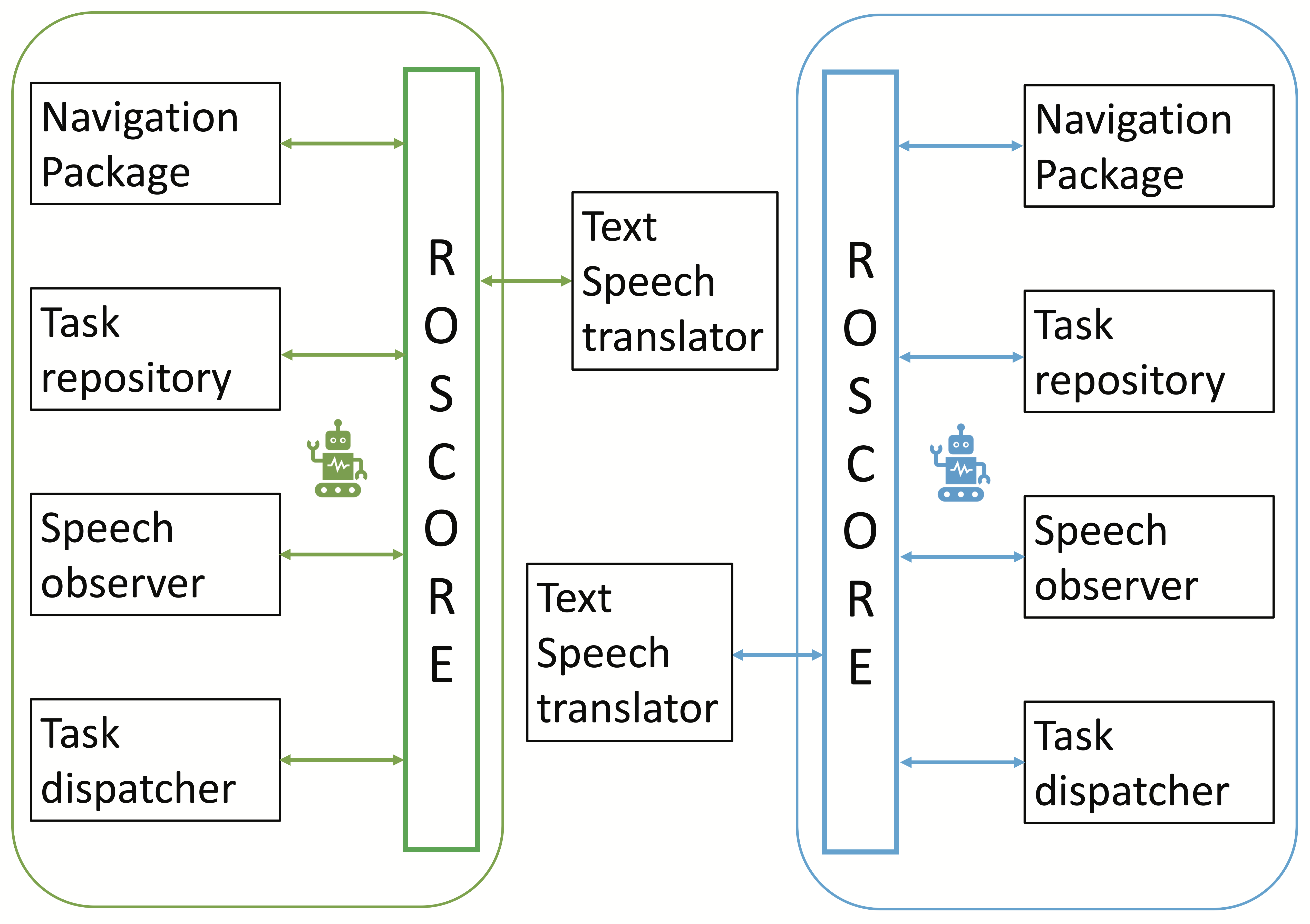}
    \caption{Illustration of two robots running in a decentralised mode for coordination and modules that are involved to achieve coordination.}
    \label{fig:tworobots}
\end{figure}

The implementation of the verbal communication platform on two robots are shown in Figure \ref{fig:tworobots}. As can be seen, the two robots are running their own roscore for verbal communication, which indicates that the proposed platform is decentralised (note traditional MRS based on Robot Operating System (ROS) usually needs only one roscore). The robots use task repository to manage tasks and update the repository upon task completion. To benefit the community, we have opensourced the codes at \href{https://github.com/jynxmagic/MSc_AI_project}{GitHub}.

\section{Results and Analysis}
\subsection{Setting Up}
An array of experiments and simulations were conducted to demonstrate the performance of the proposed verbal communication platform for MRS. To be specific, there are mainly two types:

\begin{enumerate}
    \item Assessments of the verbal communication performance of the platform.
    \item Assessments of the verbal communication platform for MRS coordination and collaboration.
\end{enumerate}

Two Turtlebot Burger robots with Raspberry Pi 3B were used for experiments. Unbuntu 20.04 LTS and ROS were used for implementation. The Picovoice is adopted for both speech-to-text and text-to-speech generation. The advantage of Picovoice is that it allows for offline translation of speech and text in a bi-direcitonal manner. Note that all the systems and software are chosen this way so that the verbal communication platform does not rely on any communication infrastructure, regardless of being wired or wireless. 

\subsubsection{Effectiveness of Wake Words}

\begin{figure}[tbp]
    \centering
    \includegraphics[width=1.0\linewidth]{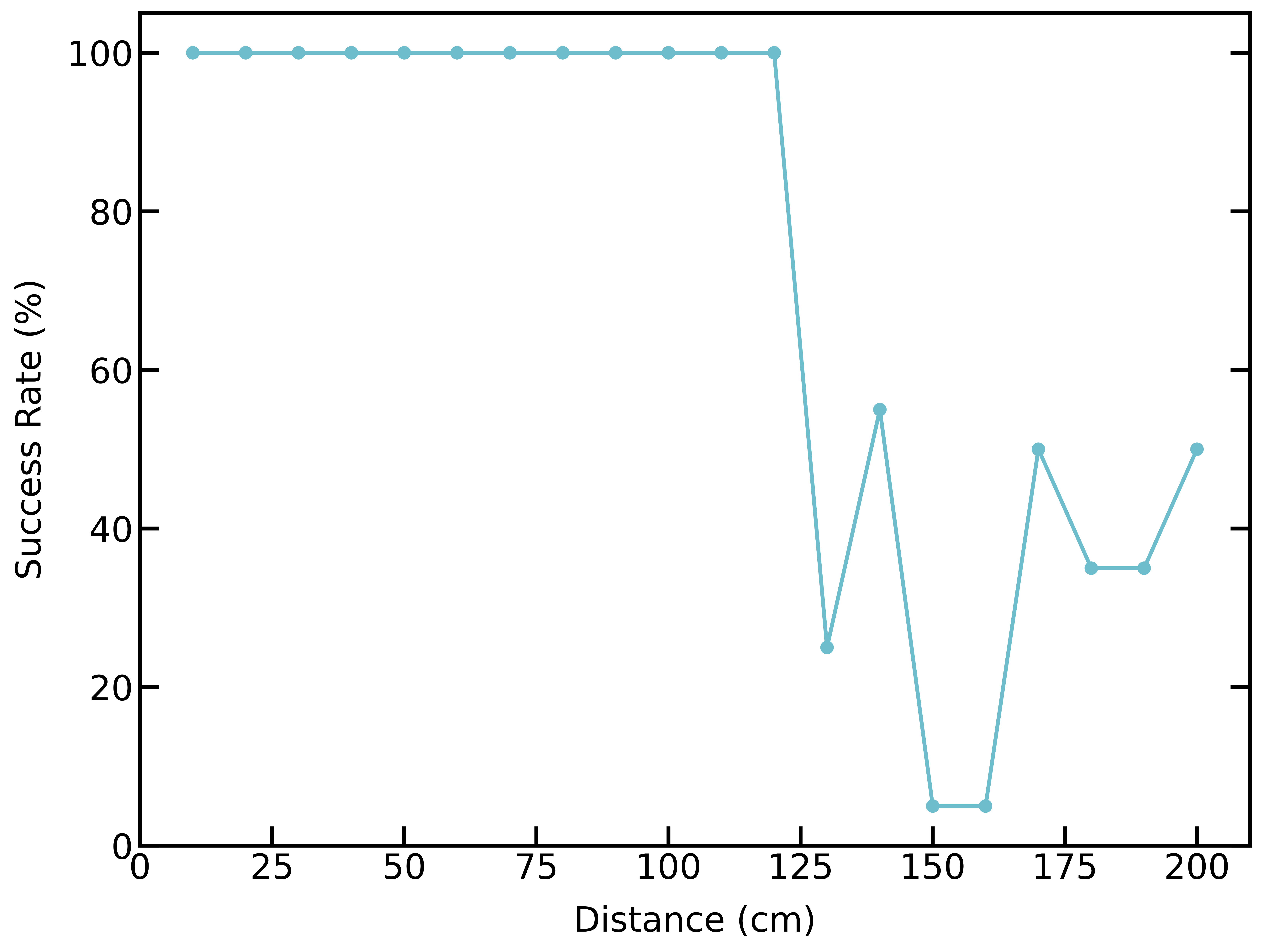}
    \caption{Success rate of the wake-word detection in different distances. The ambient noise level is 20 dB.}
    \label{fig:wake-word-fail}
\end{figure}

Wake words are used following the work pattern of Amazon Alexa and Apple Siri, etc. to invoke robots to collaborate in this paper. One critical issue is the effective distance in which the robots could be invoked and accept and execute commands conveyed by verbal signals. Two robots installed with speakers (also work as receivers) were set to speak to each other, using the wake words. The distance between the two robots was increased at a step of 10 cm, and at each distance 20 times of tests were carried out. The wake words success rate at each distance was calculated, as shown in Figure \ref{fig:wake-word-fail}. Note all the results were obtained at a 20 dB ambient noise level. One can see a sharp decrease of success rate after a certain distance, which is 120 cm in this case. For convenience, we call this distance the \textit{pivotal distance}. The existence of the pivotal distance is due to the degrading of the strength of the verbal signals, which makes it more challenging for Picovoice to translate verbal signals into texts correctly. 

\begin{figure}[tbp]
    \centering
    \includegraphics[width=1.0\linewidth]{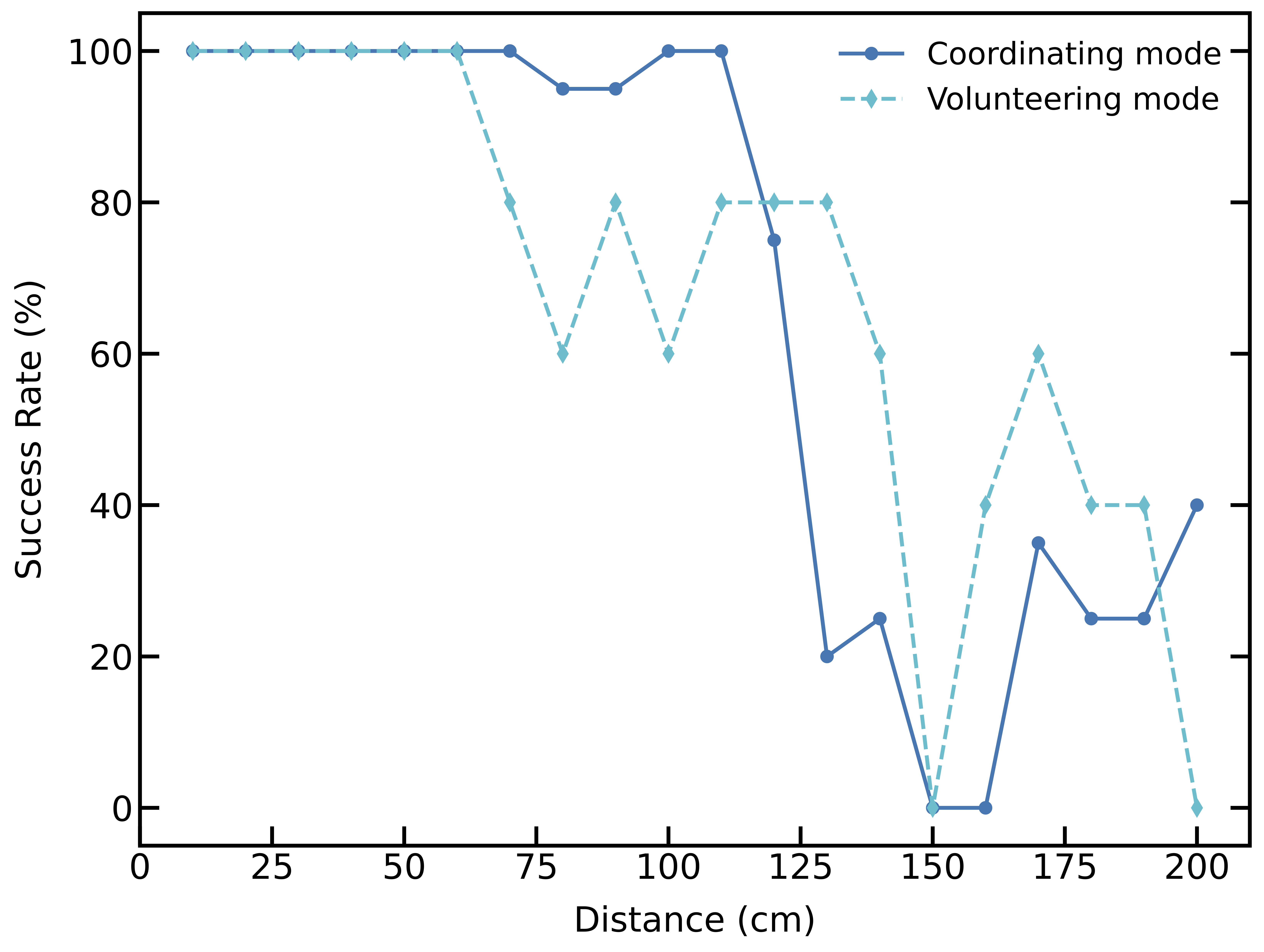}
    \caption{Success rate of the verbal communication platform in difference working modes. The ambient noise level is 20 dB.}
    \label{fig:simple_communication_simulation}
\end{figure}

\subsubsection{Multi-Robot Coordination based-on the Verbal Communication Platform}
To demonstrate that the verbal communication platform can be used by a MRS for coordination, two robots installed with the verbal communication platform were used to carry out the experiments. For clarity, two subsets of experiments were conducted depending on the working modes of the MRS.

\begin{itemize}
    \item Coordinating mode, where a robot (coordinator) asks the other robot to conduct certain tasks, such as `\textit{Hi, Rob, I am Bot, water the yellow rose}'! 
    \item Volunteering mode, where robots talk to each other to collaborate to deliver a certain task. For instance, if the task is `\textit{water plants}', then robot \textit{Rob} can volunteer to water the yellow rose and robot \textit{Bot} can then water the red rose.
\end{itemize}


\begin{figure*}[!h]
    \centering
    \subfigure[Robot \textit{Rob}'s Task Execution Trails]{
        \includegraphics[width=0.45\linewidth, height=0.42\linewidth]{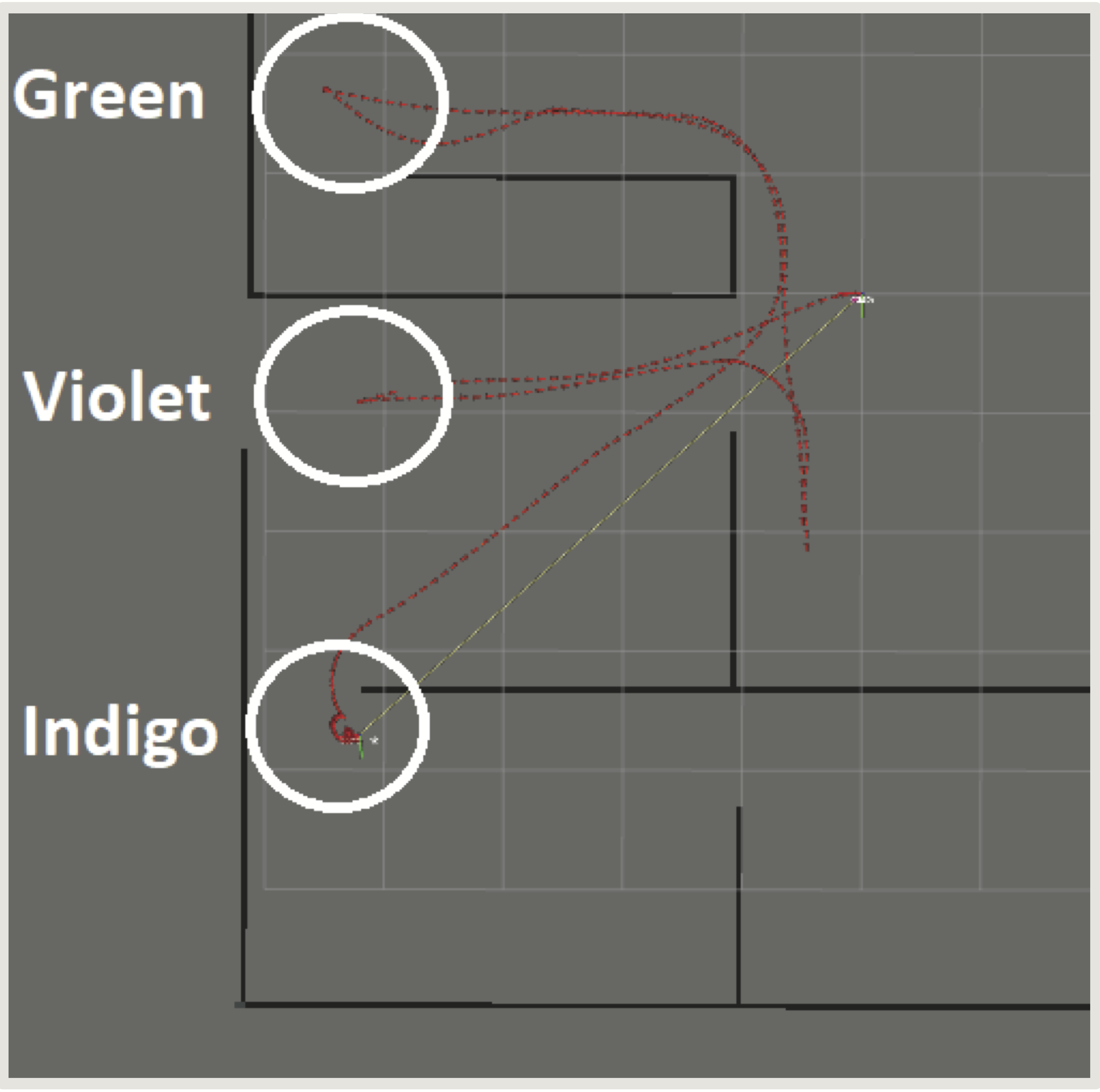}
        \label{fig:task_1}
    }
    \subfigure[Robot \textit{Bot}'s Task Execution Trails]{
	\includegraphics[width=0.45\linewidth, height=0.42\linewidth]{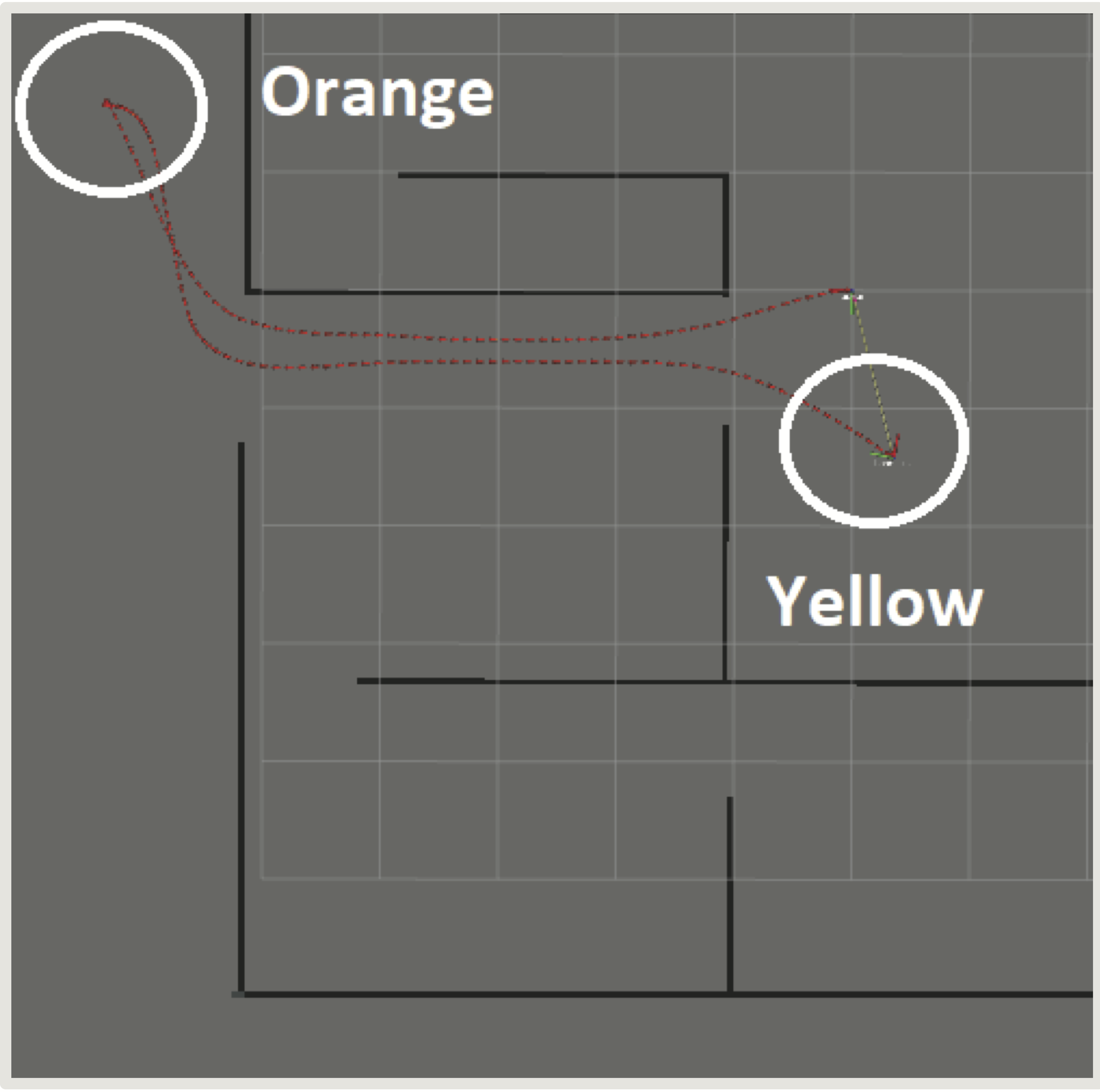}
        \label{fig:task_2}
    }
    \caption{Simulation of Two Robots, i.e., Robots \textit{Rob} and \textit{Bot} Collaborating on Watering Plants, with Each Circle Represents a Specific Plant. The Execution Trails of the Robots are Shown to Demonstrate How the Tasks Being Achieved by the Two Robots Seperately.}
    \label{fig:task}
\end{figure*}

Figure \ref{fig:simple_communication_simulation} shows the success rate of the two experiments. To be specific, in the coordinating mode, the coordinating robot asks the other robot to perform a task at various distances, with 20 times at each distance. In the volunteering mode, the two robots verbally communicate first to allocate tasks and then execute the tasks. One can see the significant decline of success rate after the pivotal distance in both cases. This is because the wake words get difficult for the robots to recognise over the pivotal distance. The success rate before hitting the pivotal distance is, however, quite high for both cases. It is worth mentioning that for the volunteering mode, the success rate starts dropping at the distance of 70 cm. This is due to the fact that after one robot such as Rob volunteers to do a task, it starts moving straight away from the other robot, and the increased distance eventually undermines the success rate.



Figure \ref{fig:task} shows the simulated collaboration of two robots, i.e., \textit{Rob} and \textit{Bot}, using the proposed verbal communication platform for the `\textit{water plants}' task. In the task, there are five plants to water, i.e., plant \textit{Green}, \textit{Violet}, \textit{Indigo}, \textit{Orange}, and \textit{Yellow}. They are registered in the task repository. After initial greetings, \textit{Rob} and \textit{Bot} will randomly pick a task for themselves from the task repository, and execute the tasks. After the successful completion of the current task, either robot will pick the next task and update the task repository. Because the shared nature of the task repository, the update will automatically notify the other robot. The process carries on until no tasks left in the task repository. As shown in Figure \ref{fig:task}, \textit{Rob} has watered three plants and \textit{Bot} has watered the left two upon completion of the \textit{water plants} task.

\subsection{Ablation}
As text-to-speech and speech-to-text carried out by Picovoice is a critical factor in determining the performance of the proposed verbal communication platform, we further investigated how it impacts the verbal communication platform. It is worth noting that Picovoice is not opensource, so all our results are generated by calling Picovoice API. 

\subsubsection{Text to Speech}

In our simulations and experiments, we noticed that most text to speech translations were correct, but there were still chances to get confusing results. In the English language, some words are spelt the same but have multiple pronunciations. These are called `homophones'. For instance, the word \textit{read} can be pronounced `r-eed', as in `comprehend the meaning of symbols', or `r-ehd' - past tense form of `r-eed'. These two words are both spelt the same, though with a different meaning and pronunciation. In every case during text to speech, the word is encoded as `r-eed', regardless of context used. This can introduce some major errors into the translation process, which are discussed some more in the speech to text section below. Table \ref{tab:my_label} gives more examples of English words and their various methods of communication, along with how Picovoice pronounces the word.

\begin{table}[tbp]
    \centering
    \caption{Possible and used pronunciations for homophones}
    \label{tab:my_label}
    \begin{tabular}{ccc}
        \hline
        \textbf{Word} & \textbf{Possible Pronunciations} & \textbf{Used Pronunciation} \\
        \hline
         read & r-ehd, r-eed & r-eed  \\
         bass & b-ass, b-ase & b-ase\\
         row & r-oh, r-ow & r-oh\\
         bow & b-ow, b-oh & b-oh\\
         tear & t-e-rh, t-ee-r & t-e-rh\\
         live & l-iv, l-eye-ve & l-eye-ve\\
         wind & w-in-d, w-eye-nd & w-in-d\\
         wound & w-oo-nd, w-ow-nd & w-oo-nd\\
         close & cl-oh-s, cl-oh-ze & cl-oh-s\\
         excuse & eh-cs-oos, eh-cs-use & eh-cs-oos \\
         polish & p-oh-l-i-sh, p-o-l-i-sh & p-o-l-i-sh\\
         lead & l-ed, l-eed & l-eed \\
         dove & d-oh-ve, d-uh-ve & d-uh-ve \\
         graduate & g-rad-u-ate, g-rad-u-at & g-rad-u-at \\
         \hline
    \end{tabular}
\end{table}
\subsubsection{Speech to Text}
The homophones issues occur in the speech to text stage as well. Using the pronunciation of `read' as an example. There are two possible pronunciations for this word. As listed in Table \ref{tab:my_label}, Picovoice uses the word as pronounced `r-eed'. In translation of the word `red' as in color, the speech to text module translated the word into `read'. The following text to speech module will use the `r-eed' pronunciation and eventually deliver `red' as in color to `read' as in `comprehend the meaning of symbols'. 

In this paper, the issue was mitigated by using alternatives of such words. For instance, when the robot failed to translate a word correctly, the phrase which triggered instead of the correct phrase was recorded. Table \ref{tab:translations_failed} shows the different phrases which triggered according to some specific keywords - the objectives. For example, a frequent error by the translation was from `indigo' to `in the go'. Should `in the go' be listed as an alternative phrase to `indigo', it would be possible to improve the robots performance drastically. With many such small improvements as this, it is possible to improve the perfect translation performance up to a distance of 354 cm with ambient noise level at 20 dB.

\begin{table}[tbp]
    \centering
    \caption{Incorrect translations of objectives in complex communications}
    \label{tab:translations_failed}
    \begin{tabular}{ccc}
        \hline
         \textbf{Word} & \textbf{Translation} & \textbf{Count}   \\
         \hline
         Indigo & in the go & 3 \\
         Indigo & him to go & 4 \\
         Indigo & do it again & 3 \\
         Indigo & you to know & 1 \\
         Yellow & you know & 2 \\
         Orange & a range & 1 \\
         \hline
    \end{tabular}
\end{table}

\section{Conclusion}
A verbal communication platform is proposed in this paper following the principles of being adaptable to environments and robot types, and transparent and secure to humans. The platform enables a multi robot system to communicate verbally and therefore less infrastructure dependent. This makes the platform promising in expanding the application of multi-robot systems into network lacking scenarios. A series of experiments were conducted to demonstrate the effectiveness of the verbal communication platform in terms of communication efficiency as well as its application in multi-robot coordination. Future works will deal with coordination in a larger number robots using the verbal communication platform.

\section*{ACKNOWLEDGMENT}

This work was funded by `Constructing accurate 3D human with deep learning in digital twin environments for safe human-robot collaboration' (Grant No. 914175).

\bibliographystyle{IEEEtran}
\bibliography{TalkingRobots}


\end{document}